\documentclass[conference]{IEEEtran}
\IEEEoverridecommandlockouts
\usepackage{cite}
\usepackage{amsmath,amssymb,amsfonts}
\usepackage{algorithmic}
\usepackage{graphicx}
\usepackage{textcomp}
\usepackage{xcolor}
\usepackage{booktabs}
\usepackage{array}
\usepackage{xcolor}
\usepackage{colortbl}
\usepackage{tabularx}
\usepackage{hyperref}

\usepackage[capitalise,nameinlink]{cleveref}  

\crefname{figure}{Fig.}{Figs.}
\crefname{table}{Tab.}{Tabs.}

\def\BibTeX{{\rm B\kern-.05em{\sc i\kern-.025em b}\kern-.08em
    T\kern-.1667em\lower.7ex\hbox{E}\kern-.125emX}}

\begin{document}

\title{SSG-Dit: A Spatial Signal Guided Framework for Controllable Video Generation\\
\thanks{
        \par\noindent\rule[0.5ex]{\linewidth}{0.4pt}\par
    This work has been submitted to the IEEE for possible publication. Copyright may be transferred without notice, after which this version may no longer be accessible.
}
}

\author{
    \IEEEauthorblockN{
        Peng Hu\IEEEauthorrefmark{2,}\IEEEauthorrefmark{1},
        Yu Gu\IEEEauthorrefmark{1},
        Liang Luo\IEEEauthorrefmark{1}, 
        and Fuji Ren\IEEEauthorrefmark{3}\IEEEauthorrefmark{1,}
    }
    
    \IEEEauthorblockA{\IEEEauthorrefmark{1}
        School of Computer Science and Engineering, University of Electronic Science and Technology of China, Chengdu, China
    }
     
}


\maketitle

\begin{abstract}
Controllable video generation aims to synthesize video content that aligns precisely with user-provided conditions, such as text descriptions and initial images. However, a significant challenge persists in this domain: existing models often struggle to maintain strong semantic consistency, frequently generating videos that deviate from the nuanced details specified in the prompts. To address this issue, we propose SSG-DiT (Spatial Signal Guided Diffusion Transformer), a novel and efficient framework for high-fidelity controllable video generation. Our approach introduces a decoupled two-stage process. The first stage, Spatial Signal Prompting, generates a spatially aware visual prompt by leveraging the rich internal representations of a pre-trained multi-modal model. This prompt, combined with the original text, forms a joint condition that is then injected into a frozen video DiT backbone via our lightweight and parameter-efficient SSG-Adapter. This unique design, featuring a dual-branch attention mechanism, allows the model to simultaneously harness its powerful generative priors while being precisely steered by external spatial signals. Extensive experiments demonstrate that SSG-DiT achieves state-of-the-art performance, outperforming existing models on multiple key metrics in the VBench benchmark, particularly in spatial relationship control and overall consistency.
\end{abstract}


\section{Introduction}
Diffusion models have recently revolutionized video generation, enabling the synthesis of high-fidelity, dynamic content \cite{wang2023videocomposer,li2024survey,singh2023survey}. A key frontier in this domain is controllable video generation\cite{wang2025survey}, where the goal is to create videos that precisely adhere to user-specified conditions. To this end, a significant body of work has focused on incorporating explicit spatial conditions, such as object trajectories \cite{huang2024vbench++,shi2024motion,qiu2024freetraj,namekata2024sg,wang2019vatex,jiang2024videobooth} or scene layouts \cite{deng2024dragvideo,chai2023stablevideo,liu2023boosting}, to provide fine-grained control over video elements. However, a critical limitation persists: while these methods excel in following explicit geometric constraints, they often fail to interpret rich, semantically rich spatial instructions embedded within natural language \cite{zhou2024survey,sun2024sora,ma2025controllable}. This leads to a "semantic drift," where a generated video might follow a trajectory but miss the abstract intent, such as a character "slowly approaching the camera." This gap arises because conventional spatial controls are treated as rigid overlays, disconnected from the deep semantic understanding of the prompt.
\begin{figure}
    \centering
    \includegraphics[width=1\linewidth]{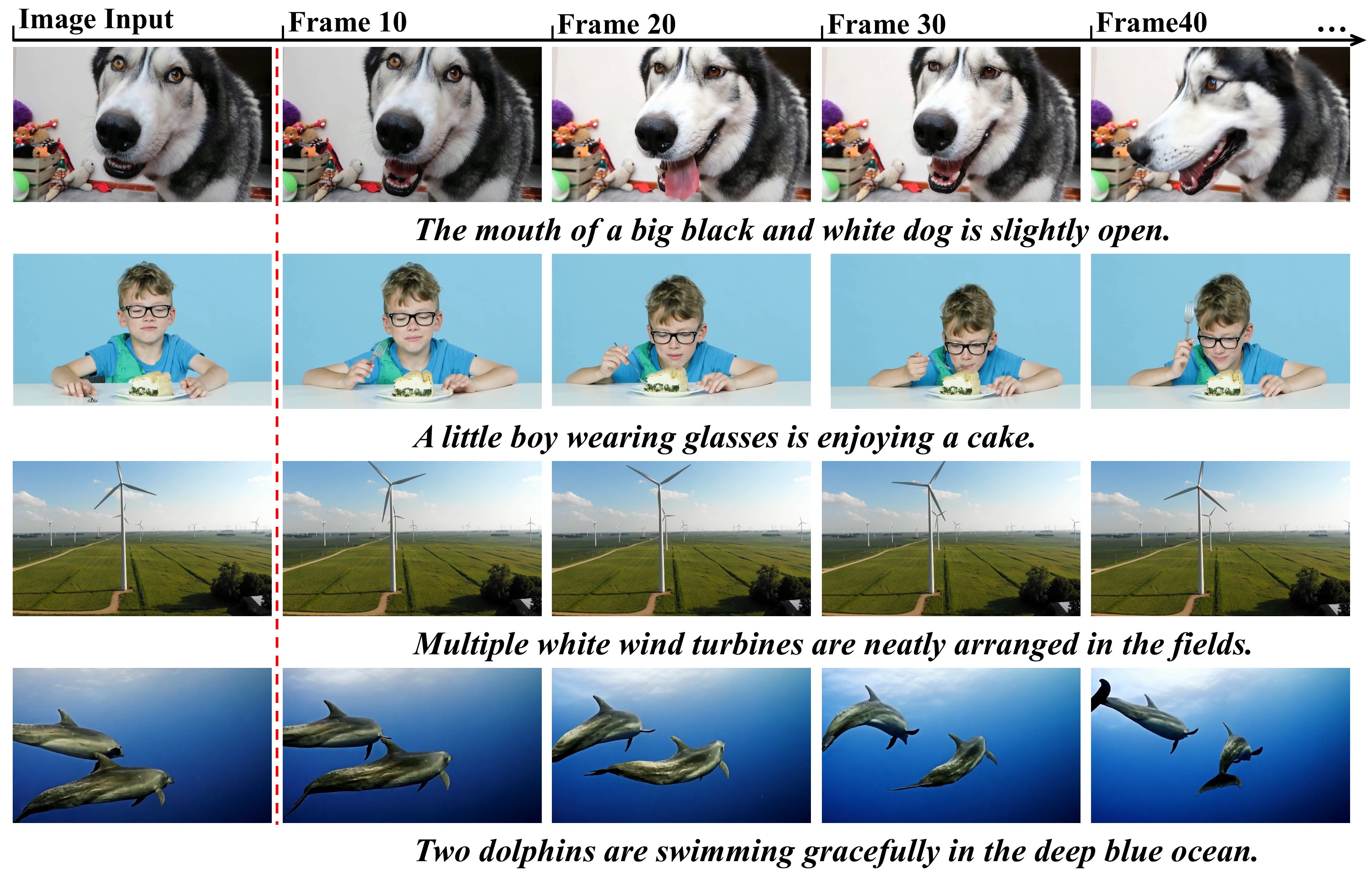}
    \caption{Qualitative results of SSG-DiT. Examples showcasing the generation of diverse and temporally coherent videos from a single image under various text prompts.}
    \label{fig:1}
\end{figure} 
\begin{figure*}[t]
    \centering
    \includegraphics[width=1\linewidth]{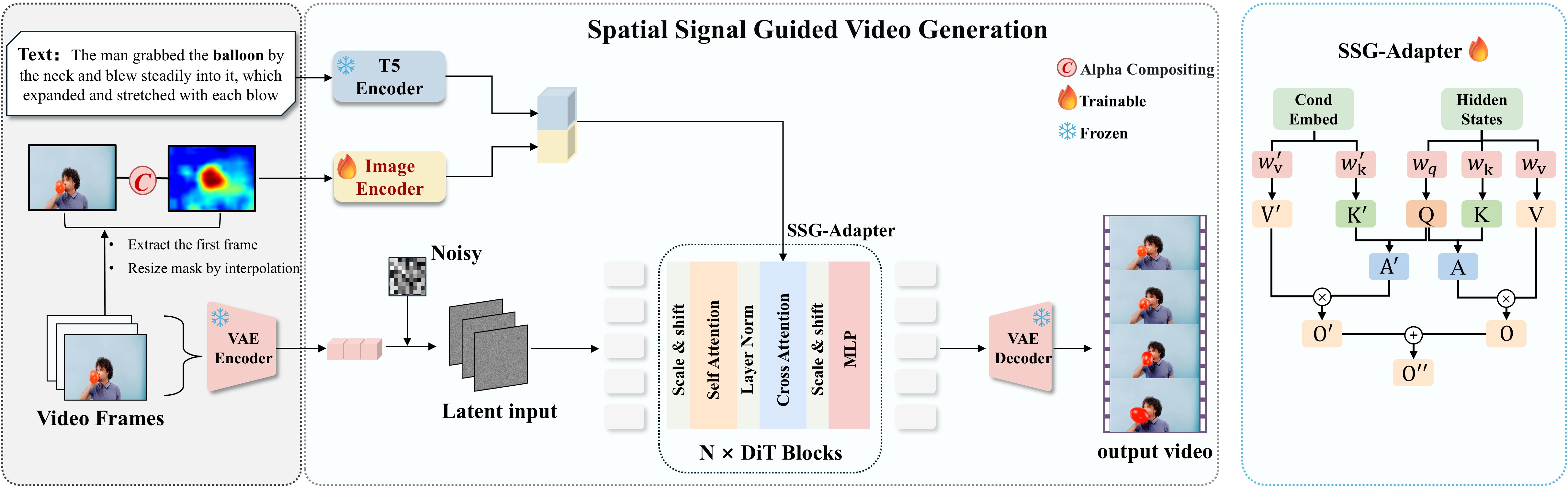}
    \caption{The architecture of our proposed SSG-DiT framework. (left) The overall pipeline, illustrating how the visual prompt and text prompt are injected into the DiT backbone via the SSG-Adapter. (right) A detailed view of the SSG-Adapter, highlighting its dual-branch attention mechanism for fusing conditional guidance with the model's hidden states.
}
    \label{fig:2}
\end{figure*}

To address this challenge, we propose SSG-DiT, a novel framework that instills semantically informed spatial control into diffusion transformers (DiT) \cite{peebles2023scalable}. Our approach features a two-stage decoupled architecture, as shown in \Cref{fig:2}. First, our Spatial Signal Prompting stage dynamically generates a text-aware visual prompt leveraging intermediate features from a pretrained CLIP model \cite{radford2021learning}. This prompt effectively translates abstract textual semantics into concrete spatial guidance. Second, following ControlNet\cite{zhang2023adding}, we introduce a lightweight, parameter-efficient SSG-Adapter that injects this visual prompt, along with the text, as a joint condition into a frozen video DiT backbone. The adapter's dual-branch attention mechanism enables the model to be meticulously guided by these rich spatial signals while preserving its powerful generative priors.

Our main contributions are: (1) We identify and tackle the problem of semantic drift for nuanced spatial instructions in video generation. (2) We propose a novel Spatial Signal Prompting mechanism to generate dynamic, text-aware visual guidance. (3) We design a parameter-efficient SSG-Adapter for effective guidance injection without full model fine-tuning. (4) We demonstrate through extensive experiments that SSG-DiT achieves state-of-the-art performance, significantly outperforming existing models in spatial control and overall consistency on the VBench benchmark.

\section{Method}

In this paper, we propose SSG-DiT, a novel framework for controllable video generation that synergizes the generative power of diffusion transformers (DiT) with precise spatial control. To address the challenge of semantic drift in video generation, our method introduces a decoupled two-stage architecture, as illustrated in \Cref{fig:2}. The first stage, Spatial Signal Prompting, leverages intermediate features from a pretrained CLIP model to dynamically generate a text-aware visual prompt that encodes spatial guidance. In the second stage, this visual prompt and the original text description form a joint multimodal condition, which is efficiently injected into a frozen video DiT backbone via our lightweight, parameter-efficient SSG-Adapter. This design enables fine-grained control over video content while preserving the model's powerful generative priors, significantly enhancing semantic consistency between the generated output and the user's prompts.



\begin{figure}[htbp]
    \centering
    \includegraphics[width=1\linewidth]{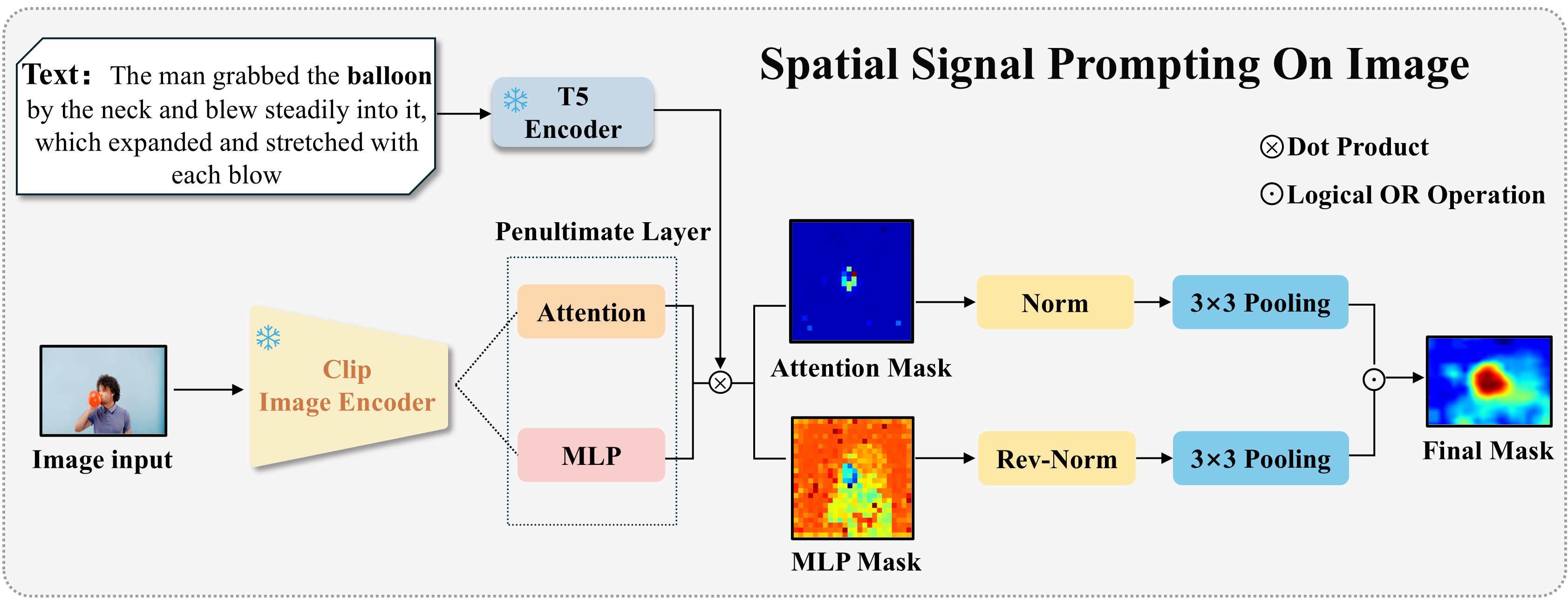}
    \caption{The pipeline for Spatial Signal Prompting. Our method generates a text-aware visual prompt by extracting and fusing complementary features (from MHSA and MLP layers) of a pre-trained CLIP model.
}
    \label{fig:3}
\end{figure}

\subsection{Spatial Signal Prompting}

This stage aims to generate a spatially aware visual prompt $I_{\text{prompt}}$ to guide the video synthesis process. Inspired by text-guided visual prompting techniques \cite{yu2024attention}, we dynamically create this prompt by leveraging the rich intermediate representations of a pre-trained CLIP (ViT-L/14) model, as detailed in \Cref{fig:3}.

\textit{1) Feature Extraction and Dual Mask Generation:} Our key insight is to fuse complementary features from both the Multi-Head Self-Attention (MHSA) and the Feed-Forward Network (FFN) modules within the penultimate Transformer block. The MHSA features capture global spatial layouts, while the FFN features encode higher-level, localized semantics. We extract these patch-level features, denoted as attention feature $A$ and MLP feature $M$ from the penultimate Transformer block of a pre-trained CLIP-ViT-L-14 model. Concurrently, the text description $T$ is encoded into an L2-normalized embedding $E_t$ using the CLIP text encoder: 

\begin{equation}
\begin{aligned}
E_t = \mathcal{L}_2\text{-norm}(\text{CLIP}_{\text{text}}(T)) 
\end{aligned}
\end{equation}

The attention and MLP response scores are then computed via a dot product with the text embedding and subsequently reshaped into 2D masks:

\begin{equation}
\begin{aligned}
M_{\text{attn}}=\text{Reshape}(A \cdot E_t)\in \mathbb{R}^{24 \times 24}\\
 M_{\text{mlp}} = \text{Reshape}(M \cdot E_t) \in \mathbb{R}^{24 \times 24}
\end{aligned}
\end{equation}

\textit{2) Mask Fusion:} To maximize the utility of these complementary masks, we devise a differentiated preprocessing strategy. The attention mask undergoes min-max normalization and contrast enhancement to sharpen the focal regions. In contrast, the MLP mask is processed with inverse normalization to highlight contextual information that is potentially overlooked by the attention mask. The masks are then passed through a 3x3 average pooling layer to suppress noise and regularize the spatial structure. Finally, the pre-processed masks $M_{\text{attn}}'$ and $M_{\text{mlp}}'$ are integrated using probabilistic OR fusion to produce a spatially smooth and semantically complete guidance mask.

\begin{equation}
\begin{aligned}
M_{\text{attn}}' &= \text{Enhance}\left( \mathcal{N}(M_{\text{attn}}) \right) \\
M_{\text{mlp}}'  &= \overline{\mathcal{N}}(M_{\text{mlp}}) \\
M_{\text{final}} &= M_{\text{attn}}' + M_{\text{mlp}}' - M_{\text{attn}}' \odot M_{\text{mlp}}'
\end{aligned}
\end{equation}

\textit{3) Image Prompt Synthesis:} To apply the guidance mask $M_{\text{final}}$ to high-resolution pixel-space synthesis, we first upscale it to the original image dimensions using bicubic interpolation, which prevents blocky artifacts. The upsampled mask is then linearly normalized to the [0, 1] range to obtain $M_{\text{norm}}$, effectively creating a smooth alpha channel. A blurred background $I_{\text{bg}}$ is generated by applying a Gaussian filter to the original image $I$. Leveraging $M_{\text{norm}}$ as an alpha channel, we achieve a precise foreground-background fusion:

\begin{equation}
\begin{aligned}
I_{\text{prompted}}(i,j) = I(i,j) \cdot M_{\text{norm}}(i,j) \\ + I_{\text{bg}}(i,j) \cdot (1 - M_{\text{norm}}(i,j))
\end{aligned}
\end{equation}

\subsection{Spatial Signal Guided Video Generation via DiT}
This stage employs a pre-trained video DiT as its backbone, guided by a joint condition comprising the image prompt $I_{\text{prompt}}$ and the text description $T$.

\textit{1) Input Representation:} Latent spatio-temporal patchification: Following the standard DiT pipeline\cite{ma2024latte}, the noisy input video $x_t $ is first mapped to a low-dimensional latent space using a fixed VAE encoder, which produces $z_t \in \mathbb{R}^{L \times H \times W \times C}$. Here, $L$, $H$, $W$ and $C$ denote the frame count, height, width, and channel dimensions of the latent representation. We then partition $z_t$ into a sequence of nonoverlapping spatiotemporal patches. These patches are flattened and linearly projected to form a token sequence $X\in \mathbb{R}^{(L \cdot N) \times D}$. To preserve positional information, we augment this sequence with learnable spatio-temporal positional encodings $P_{\text{pos}} \in \mathbb{R}^{(L \cdot N) \times D}$\cite{su2024roformer}. The final input representation is thus defined as:

\begin{equation}
\begin{aligned}
X_{\text{in}} = (\text{Flatten}(\text{Patchify}(z_t)) + P_{\text{pos}}
\end{aligned}
\end{equation}

\textit{2) Multi-modal Condition Encoding:} Our framework processes textual and visual conditions via distinct encoder pathways. The text description $T$ is encoded by a frozen T5 encoder to generate the text embedding $C_{\text{text}}$. The spatial prompt $I_{\text{prompt}}$ is encoded by a lightweight and trainable image encoder $E_{\text{Image}}$ to produce the visual embedding $C_{\text{visual}}$. Subsequently, these two embeddings are concatenated along the dimension of the sequence to form a fused multimodal condition $C_{\text{fused}} = \text{Concat}(C_{\text{text}}, C_{\text{visual}})$, providing comprehensive guidance for the generation process.

\textit{3) Spatial Signal Guided Adapter (SSG-Adapter):} The SSG-Adapter is our key innovation to enable control. It is integrated into each Transformer block of the DiT and features a parallel, dual-branch attention structure to decouple the generation and guidance tasks without modifying the pre-trained weights. Its detailed structure is depicted in the right part of \Cref{fig:2}. For an input token sequence $X_{\text{in}} \in \mathbb{R}^{(L \cdot N) \times D}$, the attention is computed as follows:
\begin{itemize}
\item Self-Attention Branch: This branch reuses the frozen self-attention module of the pre-trained DiT to model the internal spatio-temporal dependencies of the video tokens. It preserves the powerful generative priors learned from large-scale data.

\begin{equation}
\begin{aligned}
O_{\text{self}} = \text{softmax}\left(\frac{QK^T}{\sqrt{d_k}}\right)V
\end{aligned}
\end{equation}

\item Cross-Attention Branch: This is a new, trainable module that injects spatial and semantic guidance from the fused condition $C_{\text{fused}}$. It shares the query vectors $Q$ from the self-attention branch but has its own trainable key $(W_K')$ and value $( W_V')$ projection matrices.
\end{itemize}

\begin{equation}
\begin{aligned}
O_{\text{cross}} = \text{softmax}\left(\frac{QK'^T}{\sqrt{d_k}}\right)V'
\end{aligned}
\end{equation}

The outputs of both branches are fused via a residual connection with the input, completing the attention operation for the block. This dual-branch design enables the model to simultaneously leverage its internal generative priors and external conditional guidance, thus achieving precise and controllable video generation.

\begin{equation}
\begin{aligned}
O_{\text{attn}} = X_{\text{in}} + O_{\text{self}} + O_{\text{cross}}
\end{aligned}
\end{equation}

\section{Experiments}

\subsection{Experimental Setup}\label{AA}
\textbf{Implementation Details.} Our framework is based on the Wan2.1~\cite{wan2025wan} text-to-video model. Following a parameter-efficient strategy, we freeze the original DiT backbone and VAE, and exclusively fine-tune our proposed SSG-Adapter and a lightweight image encoder.

\textbf{Dataset.} We curated a high-quality dataset of 33,500 1080p text-video pairs from OpenVidHD-0.4M~\cite{nan2024openvid}. The initial frame of each video is processed by our Spatial Signal Prompting stage to generate the visual condition for fine-tuning.


\begin{table}[htbp]
\centering
\caption{Quantitative comparison on the VBench benchmark}
\label{tab:model_comparison}
\setlength{\tabcolsep}{0.7pt}
\begin{tabular}{l*{8}{c}}
\toprule
\tiny
\scriptsize\textbf{Models} & 
\textbf{\tiny\begin{tabular}[c]{@{}c@{}}Subject\\Consistency\end{tabular}} & 
\textbf{\tiny\begin{tabular}[c]{@{}c@{}}Background\\Consistency\end{tabular}} & 
\textbf{\tiny\begin{tabular}[c]{@{}c@{}}Temporal\\Flickering\end{tabular}} & 
\textbf{\tiny\begin{tabular}[c]{@{}c@{}}Aesthetic\\Quality\end{tabular}} & 
\textbf{\tiny\begin{tabular}[c]{@{}c@{}}Imaging\\Quality\end{tabular}} & 
\textbf{\tiny\begin{tabular}[c]{@{}c@{}}Spatial\\Relationship\end{tabular}} & 
\textbf{\tiny\begin{tabular}[c]{@{}c@{}}Temporal\\Style\end{tabular}} & 
\textbf{\tiny\begin{tabular}[c]{@{}c@{}}Overall\\Consistency\end{tabular}} \\
\midrule
\scriptsize
{Wan\cite{wan2025wan}} & 
95.48 & 97.26 & 98.03 & \hspace{4pt}\textbf{62.85} & \hspace{5pt}66.39 & 70.13 & 22.81 & 22.58 \\
{Hunyuan\cite{kong2024hunyuanvideo}} & 
95.84 & \textbf{97.27} & 98.33 & \hspace{4pt}59.78 & \hspace{5pt}\textbf{70.78} & 77.40 & 21.92 & 24.75 \\
{Cogvideo\cite{yang2024cogvideox}} & 
95.73 & 95.51 & 97.60 & \hspace{4pt}60.89 & \hspace{5pt}62.13 & 67.81 & 24.34 & 25.96 \\
{Step-video\cite{ma2025step}} & 
96.92 & 96.36 & \textbf{98.56} & \hspace{4pt}59.78 & \hspace{5pt}69.13 & 74.60 & 24.14 & 26.03 \\
\rowcolor{gray!20} \textbf{Ours} & 
\textbf{97.40} & {97.08} & 96.94 & \hspace{4pt}59.83 & \hspace{5pt}69.50 & \textbf{78.17} & \textbf{25.12} & \textbf{26.31} \\
\bottomrule
\end{tabular}
\end{table}



\textbf{Evaluation Metrics.} We conduct a comprehensive evaluation using the VBench benchmark \cite{huang2024vbench} to assess the overall quality of the video and the consistency of the conditions. To further quantify specific capabilities, we employ three targeted metrics \cite{wei2023elite}: (1) CLIP-Text Score for text-video semantic alignment; (2) CLIP-Image Score for general content preservation; and (3) DINO Score \cite{oquab2023dinov2} for robust subject identity consistency, leveraging its sensitivity to fine-grained intraclass details.

\begin{figure}[htbp]
    \centering
    \includegraphics[width=0.9\linewidth]{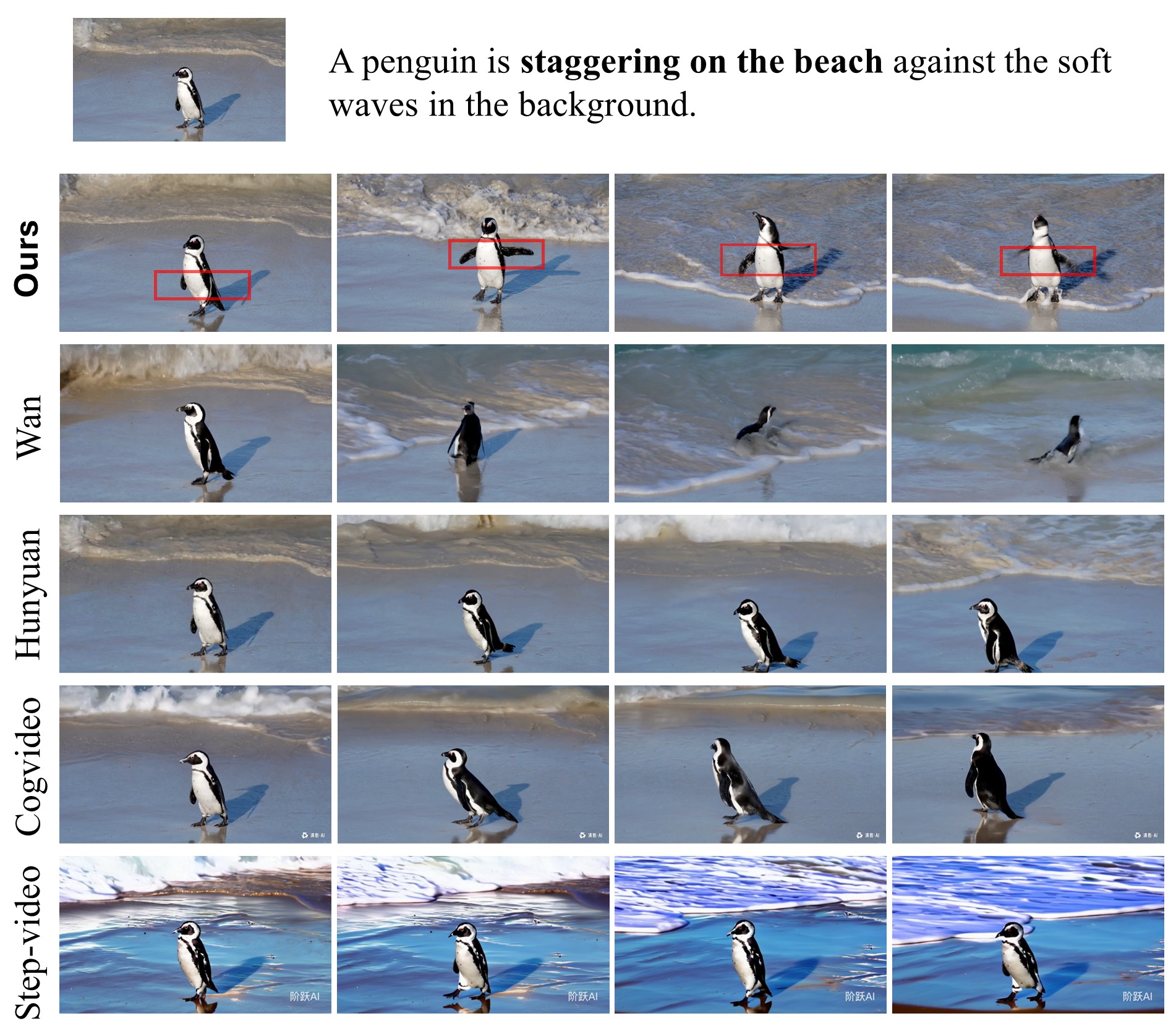}
    \caption{Qualitative comparison on motion and appearance fidelity. Our method accurately preserves the subject's appearance and generates a more vivid "staggering" motion compared to baselines.}
    \label{fig:4}
\end{figure}

\begin{figure}[htbp]
    \centering
    \includegraphics[width=0.9\linewidth]{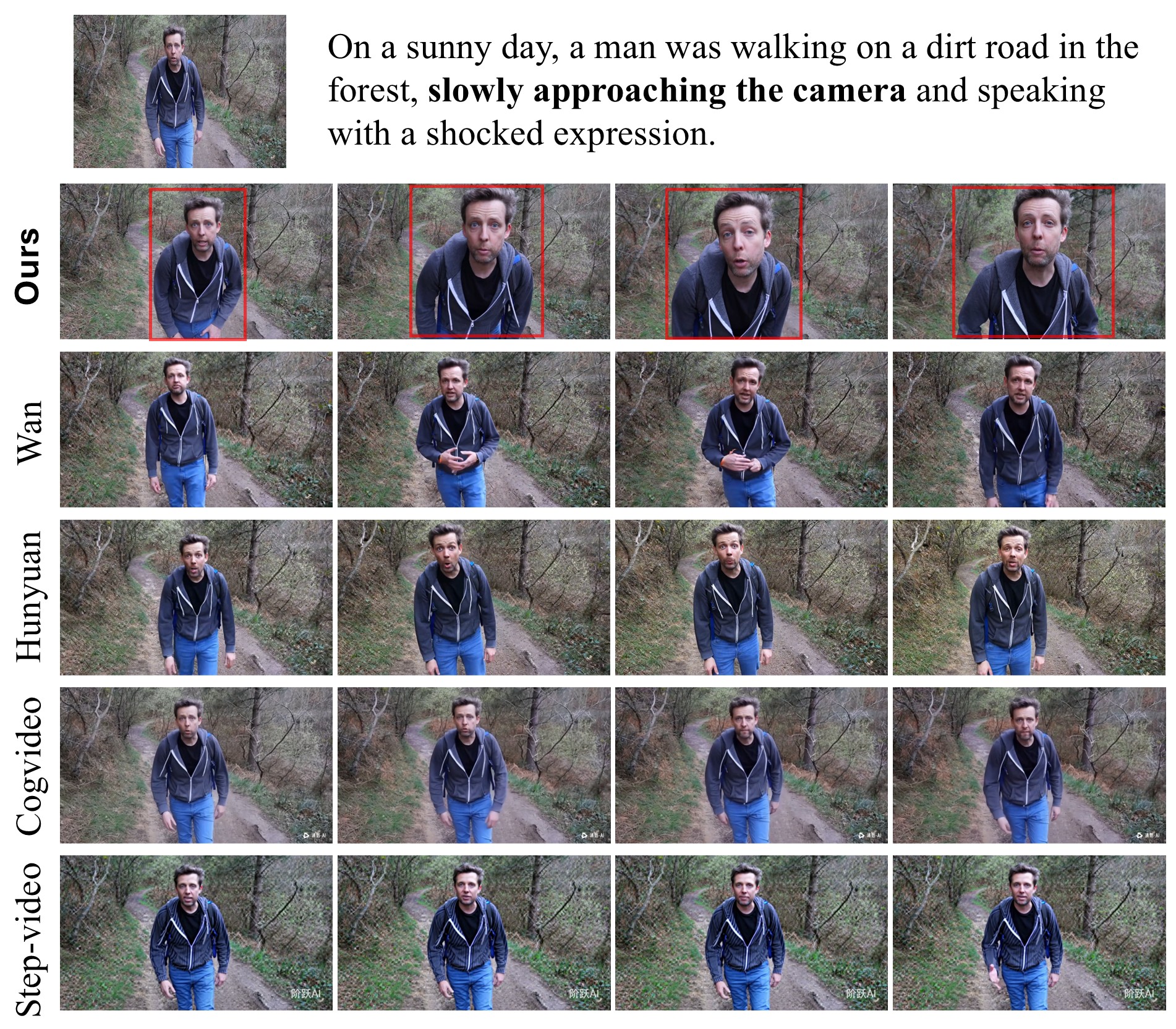}
    \caption{ Given the prompt "...slowly approaching the camera...", our method successfully generates the specified dynamic motion, while competing methods fail to capture the continuous movement.}
    \label{fig:5}
\end{figure}

\subsection{Results and Comparisons}

We quantitatively compare SSG-DiT with several SOTA models, with results presented in \Cref{tab:model_comparison} and \Cref{tab:clip_dino_scores}. On the VBench benchmark, our method achieves the highest scores in crucial consistency dimensions, including spatial relationship (78.17), temporal style (25.12), subject consistency (97.40) and overall consistency (26.31). This superior performance validates the effectiveness of our spatial signal guidance in maintaining fidelity to nuanced user prompts. The supplementary metrics in \Cref{tab:clip_dino_scores} further reinforce this conclusion. SSG-DiT consistently outperforms competitors in CLIP-Text, CLIP-Image, and DINO scores, demonstrating its exceptional ability to simultaneously preserve textual semantics and fine-grained visual identity from the conditioning prompts.

\subsection{Ablation Study}

We conducted ablation studies to validate our design choices, as shown in \Cref{tab:ablation}. Removing the entire SSG module (w/o SSG) leads to a catastrophic performance drop (e.g., Overall Consistency plummets from 26.31 to 18.91), confirming it as the cornerstone of our framework. Furthermore, ablating the Attention Mask and MLP Mask individually reveals their complementary roles: the former is critical for preserving subject identity (lower DINO score), while the latter is essential for capturing abstract semantics (lower CLIP-Text score). These results justify our approach of fusing both feature maps to form a comprehensive visual prompt.

\begin{table}[htbp]
\centering
\caption{Evaluation results using CLIP and DINO scores }
\label{tab:clip_dino_scores}
\setlength{\tabcolsep}{6pt} 
\renewcommand{\arraystretch}{1.2} 

\begin{tabular}{lccc}
\toprule
\textbf{Models} & \textbf{CLIP-Text ↑} & \textbf{CLIP-Image ↑} & \textbf{DINO ↑} \\
\midrule
Wan\cite{wan2025wan}        & 30.2784          & 75.1029          & 65.4392 \\
Hunyuan\cite{kong2024hunyuanvideo}    & 29.9630          & 73.9428          & 59.7856 \\
Cogvideo\cite{yang2024cogvideox}   & 30.1876          & 71.8834          & 54.6723 \\
Step-video\cite{ma2025step} & 30.0512          & 69.2213          & 50.3841 \\
\rowcolor{gray!20}
\textbf{Ours} & \textbf{30.2896} & \textbf{75.8307} & \textbf{67.5681} \\
\bottomrule
\end{tabular}
\end{table}

\begin{table}[htbp]
\centering
\caption{Ablation study on key components of our model}
\label{tab:ablation}
\setlength{\tabcolsep}{2pt}
\renewcommand{\arraystretch}{1.2}

\begin{tabular}{lcccc}
\toprule
\scriptsize\textbf{Variants} & 
\scriptsize\textbf{CLIP-Text ↑} & 
\scriptsize\textbf{CLIP-Image ↑} & 
\scriptsize\textbf{DINO ↑} & 
\scriptsize\textbf{\begin{tabular}[c]{@{}c@{}}Overall\\Consistency ↑\end{tabular}} \\
\midrule
w/o SSG             & 27.4112 & 66.7389 & 44.7825 & 18.9138 \\
w/o Attention Mask  & 27.5327 & 68.9043 & 48.9611 & 20.4293 \\
w/o MLP Mask        & 28.1734 & 70.2891 & 50.4187 & 21.1876 \\
\rowcolor{gray!20}
\textbf{Full Model} & \textbf{30.2896} & \textbf{75.8307} & \textbf{67.5681} & \textbf{26.3122} \\
\bottomrule
\end{tabular}
\end{table}

\section{Conclusion}

In this paper, we presented SSG-DiT, a novel framework designed to address the critical challenge of semantic drift in controllable video generation. By introducing a decoupled two-stage process, our method successfully enhances the semantic consistency between generated videos and complex multi-modal prompts. The core of our approach lies in the Spatial Signal Prompting stage, which generates a text-aware visual prompt, and a lightweight SSG-Adapter, which efficiently injects this fine-grained spatial guidance into a frozen video DiT backbone. This design allows for precise control over video content without requiring full model fine-tuning, thus preserving powerful generative priors. Our extensive quantitative and qualitative experiments validated the effectiveness of SSG-DiT, demonstrating state-of-the-art performance on the VBench benchmark and showcasing its superior ability to handle nuanced spatial and temporal instructions compared to existing methods.

\section{Acknowledgements}
This work was supported by the National Natural Science Foundation of China (No. U24A20250), the Sichuan Provincial Natural Science Foundation (Grant No. 2024NSFSC0506), the Key Project of the Sichuan Science and Technology Program (Grant No. 2024YFG0006) and the Sichuan Science and Technology Program (Grant No. 2024NSFTD0042).

\clearpage
\bibliographystyle{IEEEtran}
\bibliography{IEEEabrv,HUPENG_IEEE_ICASSP/main}

\begin{thebibliography}{10}
\providecommand{\url}[1]{#1}
\csname url@samestyle\endcsname
\providecommand{\newblock}{\relax}
\providecommand{\bibinfo}[2]{#2}
\providecommand{\BIBentrySTDinterwordspacing}{\spaceskip=0pt\relax}
\providecommand{\BIBentryALTinterwordstretchfactor}{4}
\providecommand{\BIBentryALTinterwordspacing}{\spaceskip=\fontdimen2\font plus
\BIBentryALTinterwordstretchfactor\fontdimen3\font minus \fontdimen4\font\relax}
\providecommand{\BIBforeignlanguage}[2]{{%
\expandafter\ifx\csname l@#1\endcsname\relax
\typeout{** WARNING: IEEEtran.bst: No hyphenation pattern has been}%
\typeout{** loaded for the language `#1'. Using the pattern for}%
\typeout{** the default language instead.}%
\else
\language=\csname l@#1\endcsname
\fi
#2}}
\providecommand{\BIBdecl}{\relax}
\BIBdecl

\bibitem{wang2023videocomposer}
X.~Wang \emph{et~al.}, ``Videocomposer: Compositional video synthesis with motion controllability,'' \emph{Advances in Neural Information Processing Systems}, vol.~36, pp. 7594--7611, 2023.

\bibitem{li2024survey}
C.~Li, D.~Huang, Z.~Lu, Y.~Xiao, Q.~Pei, and L.~Bai, ``A survey on long video generation: Challenges, methods, and prospects,'' \emph{arXiv preprint arXiv:2403.16407}, 2024.

\bibitem{singh2023survey}
A.~Singh, ``A survey of ai text-to-image and ai text-to-video generators,'' in \emph{2023 4th International Conference on Artificial Intelligence, Robotics and Control (AIRC)}.\hskip 1em plus 0.5em minus 0.4em\relax IEEE, 2023, pp. 32--36.

\bibitem{wang2025survey}
Y.~Wang, X.~Liu, W.~Pang, L.~Ma, S.~Yuan, P.~Debevec, and N.~Yu, ``Survey of video diffusion models: Foundations, implementations, and applications,'' \emph{arXiv preprint arXiv:2504.16081}, 2025.

\bibitem{huang2024vbench++}
Z.~Huang, F.~Zhang, X.~Xu, Y.~He, J.~Yu, Z.~Dong, Q.~Ma, N.~Chanpaisit, C.~Si, Y.~Jiang \emph{et~al.}, ``Vbench++: Comprehensive and versatile benchmark suite for video generative models,'' \emph{arXiv preprint arXiv:2411.13503}, 2024.

\bibitem{shi2024motion}
X.~Shi, Z.~Huang, F.-Y. Wang, W.~Bian, D.~Li, Y.~Zhang, M.~Zhang, K.~C. Cheung, S.~See, H.~Qin \emph{et~al.}, ``Motion-i2v: Consistent and controllable image-to-video generation with explicit motion modeling,'' in \emph{ACM SIGGRAPH 2024 Conference Papers}, 2024, pp. 1--11.

\bibitem{qiu2024freetraj}
H.~Qiu, Z.~Chen, Z.~Wang, Y.~He, M.~Xia, and Z.~Liu, ``Freetraj: Tuning-free trajectory control in video diffusion models,'' \emph{arXiv preprint arXiv:2406.16863}, 2024.

\bibitem{namekata2024sg}
K.~Namekata, S.~Bahmani, Z.~Wu, Y.~Kant, I.~Gilitschenski, and D.~B. Lindell, ``Sg-i2v: Self-guided trajectory control in image-to-video generation,'' \emph{arXiv preprint arXiv:2411.04989}, 2024.

\bibitem{wang2019vatex}
X.~Wang, J.~Wu, J.~Chen, L.~Li, Y.-F. Wang, and W.~Y. Wang, ``Vatex: A large-scale, high-quality multilingual dataset for video-and-language research,'' in \emph{Proceedings of the IEEE/CVF international conference on computer vision}, 2019, pp. 4581--4591.

\bibitem{jiang2024videobooth}
Y.~Jiang, T.~Wu, S.~Yang, C.~Si, D.~Lin, Y.~Qiao, C.~C. Loy, and Z.~Liu, ``Videobooth: Diffusion-based video generation with image prompts,'' in \emph{Proceedings of the IEEE/CVF Conference on Computer Vision and Pattern Recognition}, 2024, pp. 6689--6700.

\bibitem{deng2024dragvideo}
Y.~Deng, R.~Wang, Y.~Zhang, Y.-W. Tai, and C.-K. Tang, ``Dragvideo: Interactive drag-style video editing,'' in \emph{European Conference on Computer Vision}.\hskip 1em plus 0.5em minus 0.4em\relax Springer, 2024, pp. 183--199.

\bibitem{chai2023stablevideo}
W.~Chai, X.~Guo, G.~Wang, and Y.~Lu, ``Stablevideo: Text-driven consistency-aware diffusion video editing,'' in \emph{Proceedings of the IEEE/CVF International Conference on Computer Vision}, 2023, pp. 23\,040--23\,050.

\bibitem{liu2023boosting}
H.~Liu, T.~Wang, J.~Cao, R.~He, and J.~Tao, ``Boosting fast and high-quality speech synthesis with linear diffusion,'' \emph{arXiv preprint arXiv:2306.05708}, 2023.

\bibitem{zhou2024survey}
P.~Zhou, L.~Wang, Z.~Liu, Y.~Hao, P.~Hui, S.~Tarkoma, and J.~Kangasharju, ``A survey on generative ai and llm for video generation, understanding, and streaming,'' \emph{arXiv preprint arXiv:2404.16038}, 2024.

\bibitem{sun2024sora}
R.~Sun, Y.~Zhang, T.~Shah, J.~Sun, S.~Zhang, W.~Li, H.~Duan, B.~Wei, and R.~Ranjan, ``From sora what we can see: A survey of text-to-video generation,'' \emph{arXiv preprint arXiv:2405.10674}, 2024.

\bibitem{ma2025controllable}
Y.~Ma, K.~Feng, Z.~Hu, X.~Wang, Y.~Wang, M.~Zheng, X.~He, C.~Zhu, H.~Liu, Y.~He \emph{et~al.}, ``Controllable video generation: A survey,'' \emph{arXiv preprint arXiv:2507.16869}, 2025.

\bibitem{peebles2023scalable}
W.~Peebles and S.~Xie, ``Scalable diffusion models with transformers,'' in \emph{Proceedings of the IEEE/CVF international conference on computer vision}, 2023, pp. 4195--4205.

\bibitem{radford2021learning}
A.~Radford, J.~W. Kim, C.~Hallacy, A.~Ramesh, G.~Goh, S.~Agarwal, G.~Sastry, A.~Askell, P.~Mishkin, J.~Clark \emph{et~al.}, ``Learning transferable visual models from natural language supervision,'' in \emph{International conference on machine learning}.\hskip 1em plus 0.5em minus 0.4em\relax PmLR, 2021, pp. 8748--8763.

\bibitem{zhang2023adding}
L.~Zhang, A.~Rao, and M.~Agrawala, ``Adding conditional control to text-to-image diffusion models,'' in \emph{Proceedings of the IEEE/CVF international conference on computer vision}, 2023, pp. 3836--3847.

\bibitem{yu2024attention}
R.~Yu, W.~Yu, and X.~Wang, ``Attention prompting on image for large vision-language models,'' in \emph{European Conference on Computer Vision}.\hskip 1em plus 0.5em minus 0.4em\relax Springer, 2024, pp. 251--268.

\bibitem{ma2024latte}
X.~Ma, Y.~Wang, G.~Jia, X.~Chen, Z.~Liu, Y.-F. Li, C.~Chen, and Y.~Qiao, ``Latte: Latent diffusion transformer for video generation,'' \emph{arXiv preprint arXiv:2401.03048}, 2024.

\bibitem{su2024roformer}
J.~Su, M.~Ahmed, Y.~Lu, S.~Pan, W.~Bo, and Y.~Liu, ``Roformer: Enhanced transformer with rotary position embedding,'' \emph{Neurocomputing}, vol. 568, p. 127063, 2024.

\bibitem{wan2025wan}
T.~Wan, A.~Wang, B.~Ai, B.~Wen, C.~Mao, C.-W. Xie, D.~Chen, F.~Yu, H.~Zhao, J.~Yang \emph{et~al.}, ``Wan: Open and advanced large-scale video generative models,'' \emph{arXiv preprint arXiv:2503.20314}, 2025.

\bibitem{nan2024openvid}
K.~Nan, R.~Xie, P.~Zhou, T.~Fan, Z.~Yang, Z.~Chen, X.~Li, J.~Yang, and Y.~Tai, ``Openvid-1m: A large-scale high-quality dataset for text-to-video generation,'' \emph{arXiv preprint arXiv:2407.02371}, 2024.

\bibitem{kong2024hunyuanvideo}
W.~Kong, Q.~Tian, Z.~Zhang, R.~Min, Z.~Dai, J.~Zhou, J.~Xiong, X.~Li, B.~Wu, J.~Zhang \emph{et~al.}, ``Hunyuanvideo: A systematic framework for large video generative models,'' \emph{arXiv preprint arXiv:2412.03603}, 2024.

\bibitem{yang2024cogvideox}
Z.~Yang, J.~Teng, W.~Zheng, M.~Ding, S.~Huang, J.~Xu, Y.~Yang, W.~Hong, X.~Zhang, G.~Feng \emph{et~al.}, ``Cogvideox: Text-to-video diffusion models with an expert transformer,'' \emph{arXiv preprint arXiv:2408.06072}, 2024.

\bibitem{ma2025step}
G.~Ma, H.~Huang, K.~Yan, L.~Chen, N.~Duan, S.~Yin, C.~Wan, R.~Ming, X.~Song, X.~Chen \emph{et~al.}, ``Step-video-t2v technical report: The practice, challenges, and future of video foundation model,'' \emph{arXiv preprint arXiv:2502.10248}, 2025.

\bibitem{huang2024vbench}
Z.~Huang, Y.~He, J.~Yu, F.~Zhang, C.~Si, Y.~Jiang, Y.~Zhang, T.~Wu, Q.~Jin, N.~Chanpaisit \emph{et~al.}, ``Vbench: Comprehensive benchmark suite for video generative models,'' in \emph{Proceedings of the IEEE/CVF Conference on Computer Vision and Pattern Recognition}, 2024, pp. 21\,807--21\,818.

\bibitem{wei2023elite}
Y.~Wei, Y.~Zhang, Z.~Ji, J.~Bai, L.~Zhang, and W.~Zuo, ``Elite: Encoding visual concepts into textual embeddings for customized text-to-image generation,'' in \emph{Proceedings of the IEEE/CVF International Conference on Computer Vision}, 2023, pp. 15\,943--15\,953.

\bibitem{oquab2023dinov2}
M.~Oquab, T.~Darcet, T.~Moutakanni, H.~Vo, M.~Szafraniec, V.~Khalidov, P.~Fernandez, D.~Haziza, F.~Massa, A.~El-Nouby \emph{et~al.}, ``Dinov2: Learning robust visual features without supervision,'' \emph{arXiv preprint arXiv:2304.07193}, 2023.

\end{thebibliography}

\end{document}